\documentclass[10pt,twocolumn,letterpaper]{article}

\usepackage[pagenumbers]{cvpr} % To force page numbers, e.g. for an arXiv version

\usepackage{graphicx}
\usepackage{amsmath}
\usepackage{amssymb}
\usepackage{booktabs}
\usepackage{tabularray}
\usepackage{float} 
\usepackage{stfloats}
\usepackage{amsfonts,amssymb}
\usepackage{emptypage}
\usepackage[ruled]{algorithm2e}
                              
\SetCommentSty{mycommfont}

\usepackage[pagebackref,breaklinks,colorlinks]{hyperref}

\usepackage[capitalize]{cleveref}
\crefname{section}{Sec.}{Secs.}
\Crefname{section}{Section}{Sections}
\Crefname{table}{Table}{Tables}
\crefname{table}{Tab.}{Tabs.}

\def\cvprPaperID{*****} % *** Enter the CVPR Paper ID here
\def\confName{CVPR}
\def\confYear{2022}

\begin{document}

%%%%%%%%% TITLE - PLEASE UPDATE
\title{IIDM: Image-to-Image Diffusion Model for Semantic Image Synthesis}

\author{Feng Liu\textsuperscript{1}\quad Xiaobin Chang\textsuperscript{1,2}\thanks{Corresponding author}\\
\textsuperscript{1}School of Artificial Intelligence, Sun Yat-sen University, China\\
\textsuperscript{2}Key Laboratory of Machine Intelligence and Advanced Computing, China\\
{\tt\small liuf275@mail2.sysu.edu.cn} \quad
{\tt\small changxb3@mail.sysu.edu.cn}
% For a paper whose authors are all at the same institution,
% omit the following lines up until the closing ``}''.
% Additional authors and addresses can be added with ``\and'',
% just like the second author.
% To save space, use either the email address or home page, not both
% \and
% Xiaobin Chang\\
% School of Artificial Intelligence, Sun Yat-sen University\\
% Zhuhai 519082, China\\
% \\
% Ministry of Education, China\\
% {\tt\small changxb3@mail.sysu.edu.cn}
}
\maketitle

%%%%%%%%% ABSTRACT
\begin{abstract}
    Semantic image synthesis aims to generate high-quality images given  semantic conditions, i.e.\  segmentation masks and style reference images.
    Existing methods widely adopt generative adversarial networks (GANs). 
    GANs take all conditional inputs and directly  synthesize images in a single forward step.
    In this paper, semantic image synthesis is treated as an image denoising
    task and is handled with a novel image-to-image diffusion model (IIDM). 
    Specifically, the style reference is first contaminated with random noise and then progressively denoised by IIDM, guided by segmentation masks.
    Moreover, three techniques, refinement, color-transfer and model ensembles, are proposed to further boost the generation quality. They are plug-in inference modules and do not require additional training.
    Extensive experiments show that our IIDM outperforms existing state-of-the-art methods by clear margins. Further analysis is provided via detailed demonstrations.
    We have implemented IIDM based on the Jittor framework;
    code is available at \url{https://github.com/ader47/jittor-jieke-semantic_images_synthesis}.
\end{abstract}

%%%%%%%%% BODY TEXT
\section{Introduction}
\begin{figure}[t]
    \centering
    \includegraphics[width=\linewidth]{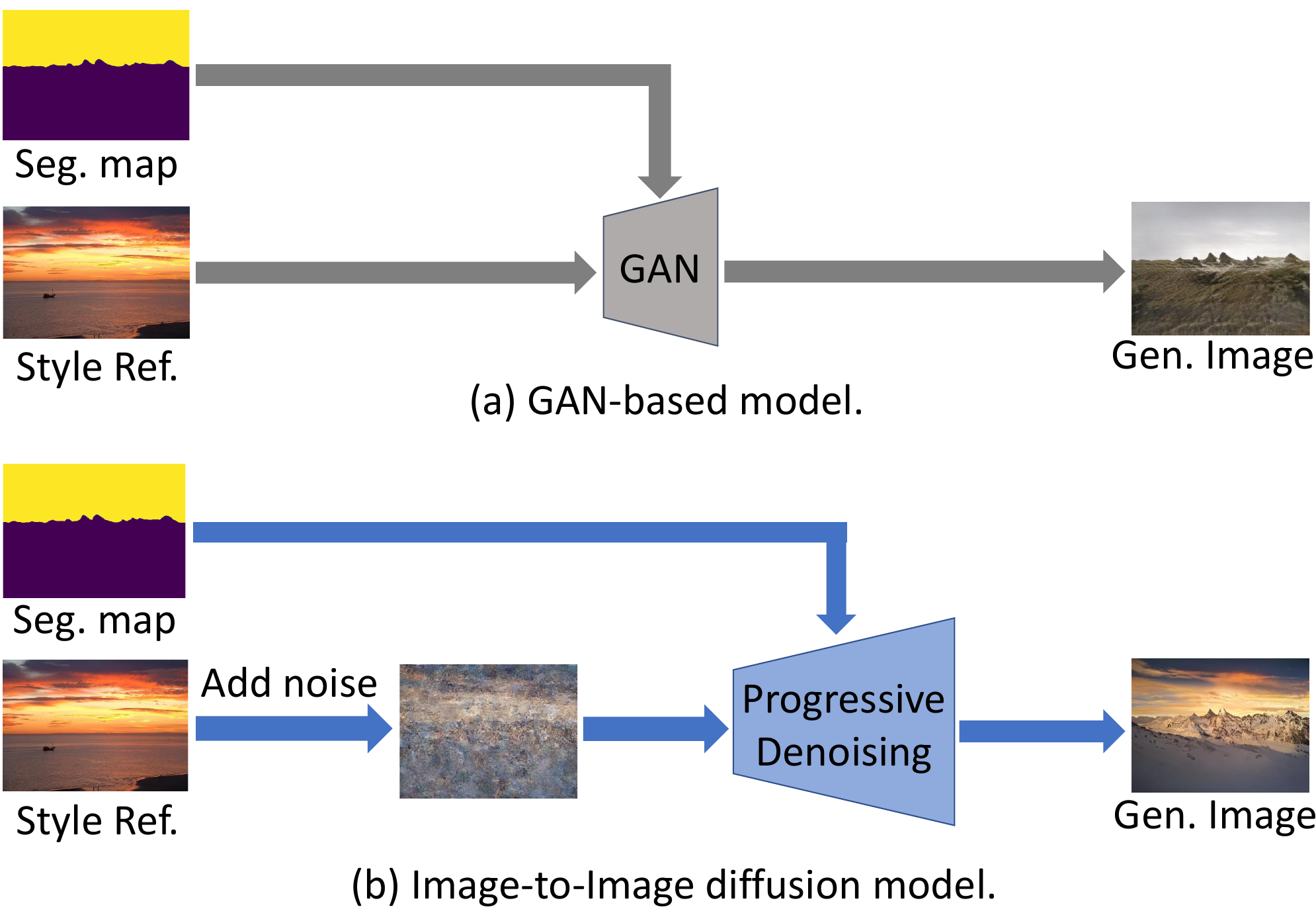}
    \caption{
    Semantic image synthesis approaches: (a) GAN-based models generate an image that simultaneously satisfies both style and segmentation conditions in a single step;
    (b) our proposed IIDM is a progressive generation (denoising) process. Different conditions are into consideration at different stages of the process.}
    \label{fig:GAN&Diffusion}
\end{figure}

Semantic image synthesis aims to generate realistic images given semantic conditions, i.e.\ segmentation masks and style reference images.
GANs~\cite{goodfellow2020generative} are widely used by existing semantic image synthesis methods, e.g.\ GauGAN~\cite{park2019semantic}, CLADE~\cite{tan2021efficient}, SEAN~\cite{Zhu_2020_CVPR}, SCGAN~\cite{wang2021image}.
Specifically, spatially adaptive normalized activation (SPADE) is learned from the semantic layouts to prevent semantic information from being forgotten,  in GauGAN.
Class-adaptive normalization (CLADE) was proposed to reduce the computational and memory load of SPADE.
Another simple and effective building block for GANs, for segmentation mask conditions, was presented in SEAN.
SCGAN is a dynamic weighted network with spatially conditional operations such as convolution and normalization. The dynamic architecture strengthens  semantic relevance and detail in synthesized images.
However, GAN-based methods simultaneously take different conditions as input and  synthesize an image through a forward step, as shown in Figure~\ref{fig:GAN&Diffusion}(a).
It is challenging for GAN-based methods to generate images that satisfy both the style and segmentation conditions in a single pass.
Moreover, GAN-based methods may lack the flexibility to handle different conditions and suffer from instability of adversarial training.

The diffusion model (DM)~\cite{dhariwal2021diffusion} is a new class of generative models and has shown great potential for image synthesis.
Unlike GAN-based models that generate images in a single step, DMs generate samples via a progressive denoising process~\cite{NEURIPS20204c5bcfec}.
However, using  style reference images with a basic DM is not straightforward. 
On the one hand, the diffusion process of basic DMs starts from  random Gaussian noise that is independent of the conditions.
On the other hand, like segmentation masks, such images can be encoded by another module but the model size will  significantly increase.

% efficiently
In this paper, we propose a novel image-to-image diffusion model which treats semantic image synthesis as an image-denoising task. To efficiently incorporate the style conditions into the diffusion process, the noise-contaminated style reference serves as a starting point.
Under the guidance of the segmentation mask, more accurate image content is generated in the given style by progressive denoising, as illustrated in Figure~\ref{fig:GAN&Diffusion}(b).
Moreover, three techniques, refinement, color-transfer and model ensembles, are applied during model inferencing to further boost  synthesis performance, especially in terms of image quality and style similarity.
Our main contributions are thus in summary:
\begin{itemize}
\item a novel diffusion model, IIDM, to efficiently and effectively incorporate different conditions on semantic image synthesis, and
\item  exploitation of three further DM inference techniques  to improve the synthesis results at a relatively low cost.
\end{itemize}

Experimental results show that IIDM outperforms existing methods by clear margins.
Implementing our IIDM with Jittor~\cite{hu2020jittor,zhou2021jittor},
we achieved first place in the semantic image synthesis track of the Third Jittor Artificial Intelligence Challenge.

%-------------------------------------------------------------------------

\section{Image-to-Image Diffusion Model}
\subsection{Semantic Image Synthesis}
Let $\mathbf{m} \in \mathbb{L}^{H\times W}$ indicate a semantic segmentation mask where $\mathbb{L}$ is a set of integers denoting  semantic labels; $H$ and $W$ denote the image height and width. Each entry in $\mathbf{m}$ provides the semantic label of a pixel. 
\(\mathbf{X}_{R}\) indicates the style reference image.
Semantic image synthesis aims to obtain a high-quality image $\mathbf{X}_G$ which follows not only the semantic segmentation mask $\mathbf{m}$ but also the style of \(\mathbf{X}_{R}\).

\subsection{Image-to-Image Diffusion Model}
\begin{figure}[t]
    \centering
    \includegraphics[width=\linewidth]{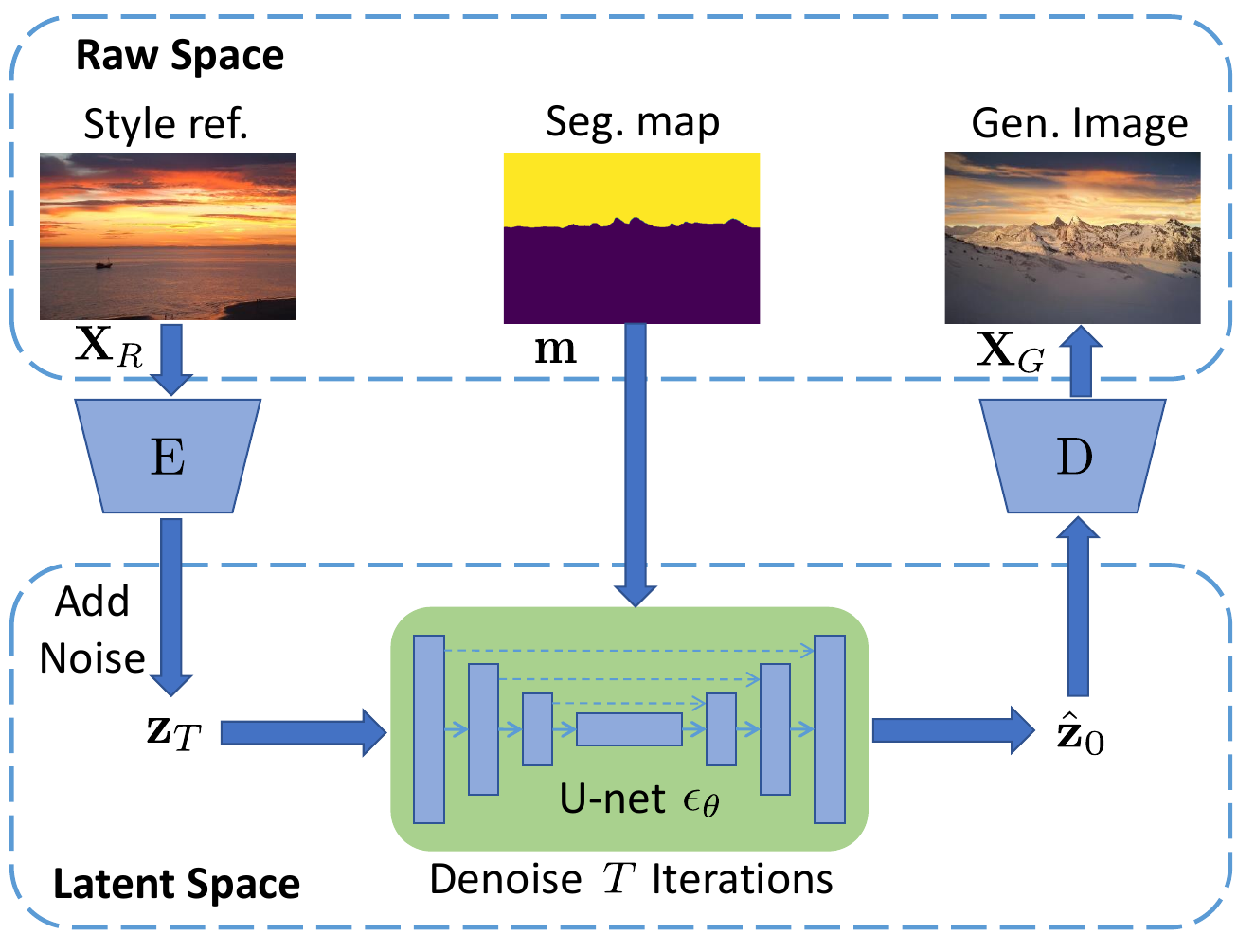}
    \caption{IIDM.
    The diffusion and denoising processes take place in the latent space via the encoder $\operatorname{E}$.
    The diffusion process first incorporates the style reference $\textbf{X}_R$ into $\textbf{z}_T$.
    The denoising process then recovers a denoised latent representation $\hat{\textbf{z}}_0$ conditioning on the segmentation map $\textbf{m}$.
    The generated image $\textbf{X}_G$ is decoded by $\operatorname{D}$ from $\hat{\textbf{z}}_0$.
    }
    \label{fig:backbone}
\end{figure}

Instead of directly denoising in the image space, our IIDM employs the latent diffusion model~\cite{rombach2021highresolution} architecture for greater efficiency.
Specifically, a pretrained autoencoder, consisting of an encoder $\operatorname{E}$ and a decoder $\operatorname{D}$, 
is used to transform the style reference image $\textbf{X}_R$ into a latent feature $\textbf{z}_0=\operatorname{E}(\textbf{X}_R)$.
The diffusion process then takes place in the latent space.
Finally, the decoder $\operatorname{D}$ projects the denoised latent feature $\hat{\textbf{z}}_0$ back into an image $\textbf{X}_G = \operatorname{D}(\hat{\textbf{z}}_0)$.
The above process is depicted in Figure~\ref{fig:backbone}.

IIDM elegantly incorporates the style information of $\textbf{X}_R$ without introducing additional modules.
Specifically, the IIDM denoising process starts from $\textbf{z}_T$ using:
\begin{equation}
    \textbf{z}_T = \sqrt{\overline{\alpha}_T}\operatorname{E}(\textbf{X}_R) + \sqrt{1-\overline{\alpha}_T} \boldsymbol{\epsilon}, 
    \label{eq:z_T}
\end{equation}
where $\beta_t = 0.001 t$, $\alpha_T =  1-\beta_T$, $T=320$ and $\boldsymbol{\epsilon} \sim \textbf{N}(\textbf{0}, \textbf{I})$ in this work.
The corresponding image of $\textbf{z}_T$ can be calculated using $\hat{\textbf{X}}_T = \operatorname{D}(\textbf{z}_T)$.
$\hat{\textbf{X}}_T$ is a noisy image with its style information kept. Therefore, this image generation process can also be treated as an \emph{image-to-image translation} from $\hat{\textbf{X}}_T$ to $\mathbf{X}_G$.
In contrast, the basic DM starts from  pure Gaussian noise without style information.

During training,
a sequence of $\textbf{z}_t$ can be obtained in the forward diffusion process using:
\begin{equation}
    % \textbf{z}_{t} = \sqrt{1-\beta_t} \textbf{z}_{t-1} + \sqrt{\beta_t} \boldsymbol{\epsilon},
    \textbf{z}_t = \sqrt{\overline{\alpha}_t}\textbf{z}_0 + \sqrt{1-\overline{\alpha}_t}\boldsymbol{\epsilon}, 
    \label{eq:forward}
\end{equation}
where $\textbf{z}_{0}=\operatorname{E}(\textbf{X}_R)$, $\boldsymbol{\epsilon} \sim \textbf{N}(\textbf{0}$, \textbf{I}), $0 \leq t \leq 1000$.
Eqn. (\ref{eq:forward}) and  (\ref{eq:z_T}) are consistent.
The reverse diffusion process $p_{\boldsymbol{\theta}}(\mathbf{z}_{0:T} | \mathbf{m})$, $T=1000$ is defined as a Markov chain with a starting point $\textbf{z}_{T}$ and ends with a latent feature $\textbf{z}_0$.
$\boldsymbol{\theta}$ indicates the model parameters of the denoising U-Net $\boldsymbol{\epsilon}_{\boldsymbol{\theta}}$.
More importantly, the progressive denoising process is guided by the segmentation mask $\textbf{m}$ so that more accurate semantic visual content is generated.
$p_{\boldsymbol{\theta}} (\mathbf{z}_{0:T} | \mathbf{m})$ is decomposed as:
\begin{equation}
    p_{\boldsymbol{\theta}}(\mathbf{z}_{0:T}|m) = p(\mathbf{z}_{T}) \prod_{t=1}^{T} p_{\boldsymbol{\theta}}\mathbf{z}_{t-1}|\mathbf{z}_t,\mathbf{m}).
    \label{eq:reverse}
\end{equation}
Given $\textbf{z}_t$, 
the diffusion model is trained to optimize the upper variational bound on negative log-likelihood (Equation~(\ref{eq:reverse})) and the learning objective becomes
\begin{equation}
    \mathcal{L}_{t-1} = \mathbb{E}_{\mathbf{z}_0, \boldsymbol{\epsilon}}[\gamma_t||\boldsymbol{\epsilon}-\boldsymbol{\epsilon}_{\boldsymbol{\theta}}(\mathbf{z}_t, \mathbf{m}, t)||^{2}],
    \label{eq:diffusion_loss}
\end{equation}
where $\boldsymbol{\epsilon} \sim \textbf{N}(\textbf{0}, \textbf{I})$ and $\mathcal{L}_{t-1}$ is the loss function at $t-1$. $\gamma_t$ is a constant at $t$.
The training procedure is depicted in Algorithm \ref{algthm:training}.

\begin{figure}[t]
    \centering
    \includegraphics[width=\linewidth]{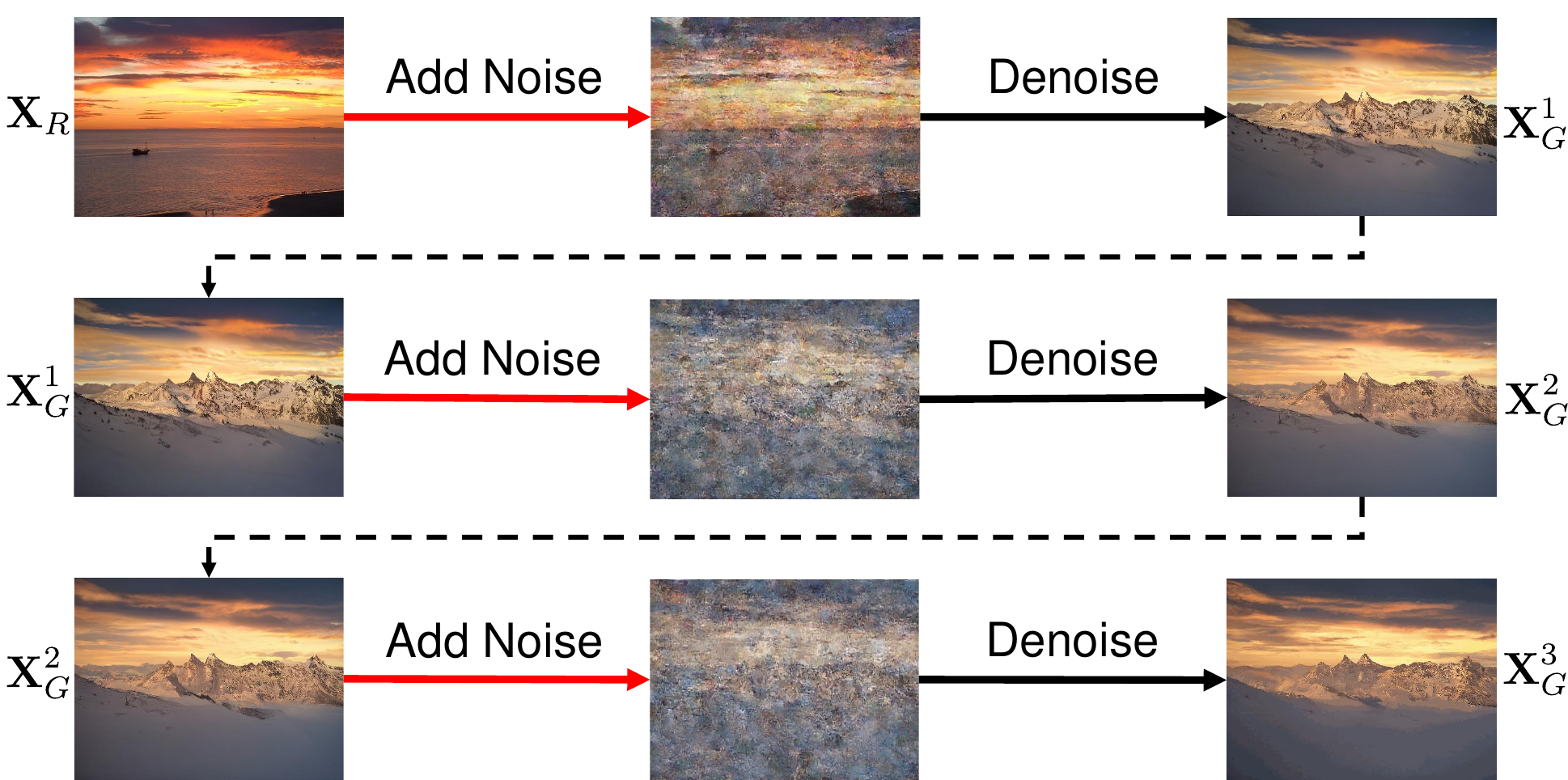}
    \caption{
    Refinement improves the quality of generated images through iterative generation. The inference
procedure can be repeated based on the synthesized image $\textbf{X}_G^k$}. 
    \label{fig:refinement}
\end{figure}

% 添加伪代码
\begin{algorithm}[t]
    % \SetAlgoNoLine
    \SetAlgoLined
    \caption{Model Training}
    \label{algthm:training}
        \KwIn{Training data $\mathcal{D}$; Image Encoder $\operatorname{E}$; 
        % Denoising 
        U-Net $\boldsymbol{\epsilon}_{\boldsymbol{\theta}}$.}
        \KwOut{Optimized $\boldsymbol{\theta}$.}
        \Repeat{converged}{
        Sample synthesis conditions $(\textbf{X}_R,\textbf{m})$ from $\mathcal{D}$;
        
        $\textbf{z}_0 = \operatorname{E}(\textbf{X}_R)$;

        $t \sim $Uniform$({1,\dots,1000})$;
        
        $\textbf{z}_t = \sqrt{\overline{\alpha}_t }{\textbf{z}}_0 + \sqrt{1-\overline{\alpha}_t}\boldsymbol{\epsilon}, \boldsymbol{\epsilon} \sim \textbf{N (\textbf{0}, \textbf{I})}$; \em\em
        \tcp{Eq.~(\ref{eq:forward})}

        $\boldsymbol{\theta} := \boldsymbol{\theta} - \nabla_{\boldsymbol{\theta}}\mathcal{L}_{t-1}$;
        \quad\tcp{Loss $\mathcal{L}_{t-1}$ (Eq.(\ref{eq:diffusion_loss}))}
        }
\end{algorithm}

\subsection{IIDM Inferencing}

Once the parameter $\boldsymbol{\theta}^*$ of the optimized diffusion model has been obtained, an image $\textbf{X}_G$ can be synthesized by progressive denoising from $t=T$ to $t=1$ using:
\begin{equation}
\left\{
\begin{aligned}
 \hat{\mathbf{z}}_{t-1} &= \tilde{\gamma}_t\hat{\mathbf{z}}_t - \tilde{\beta}_t\boldsymbol{\epsilon}_{\boldsymbol{\theta}^*}(\hat{\mathbf{z}}_t, \mathbf{m}, t)+\Tilde{\sigma}_t\boldsymbol{\epsilon}, \\
\tilde{\gamma}_t  &= \frac{\sqrt{\Bar{\alpha}_{t-1}}\beta_t}{\sqrt{\Bar{\alpha}_t}(1-\Bar{\alpha}_t)+\frac{\sqrt{\alpha}_t(1-\Bar{\alpha}_{t-1})}{1-\Bar{\alpha}_t}}, \\
 \tilde{\beta}_t &= \frac{\sqrt{\Bar{\alpha}_{t-1}}\sqrt{1-\alpha_t}\beta_t}{\sqrt{\Bar{\alpha}_t}(1-\Bar{\alpha}_t)},
\end{aligned}
\right.
\label{eq:infer}
\end{equation}
where 
$\tilde{\sigma}_t =({1-\Bar{\alpha}_{t-1}})/({1-\Bar{\alpha}_t}\beta_t)$, $\Bar{\alpha}_t = \prod_{s=1}^t\alpha_s$ and $\boldsymbol{\epsilon} \sim \textbf{N}(\mathbf{0}, \mathbf{I})$.
$\textbf{z}_{T}$ is initialized with a style reference $\textbf{X}_R$ as in Equation~(\ref{eq:z_T}). Finally, $\textbf{X}_G = \operatorname{D}(\hat{\mathbf{z}}_{0})$.

\begin{figure}[t]
    \centering
    \includegraphics[width=\linewidth]{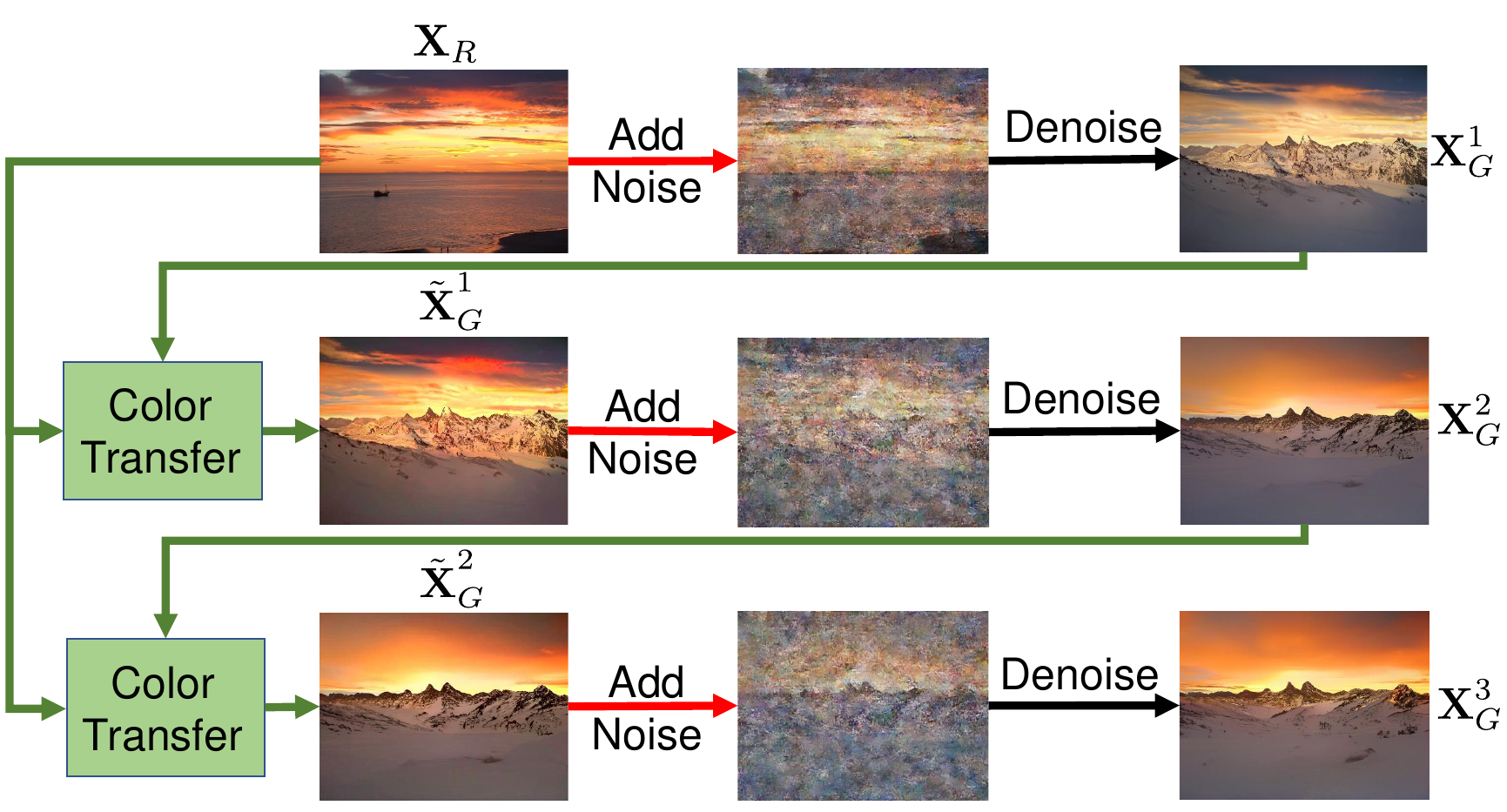}
    \caption{
    Color-transfer compensates for style information during the refinement process. At the beginning of a refinement round, style reference $\textbf{X}_R$ is incorporated by color-transfer.
    }
    \label{fig:refinement-and-color-transfer}
\end{figure}

\begin{algorithm}[!t]
    \SetAlgoLined
    \caption{IIDM Inference}
    \label{algthm:infer}
        \KwIn{
        Denoising start point $T$; %\\
        Conditions $(\textbf{X}_R, \textbf{m})$; %\\
        Autoencoder \{$\operatorname{E}$, $\operatorname{D}$\};
        Refinement iteration count $K$; %\\
        Model ensemble $\boldsymbol{\theta}^{*}$ (Eq. (\ref{eq:model_ensemble})).}
        \KwOut{Generated image $\textbf{X}_G^K$.}
        \For(\tcp*[h]{$K$ rounds  of refinement}){$k=1$ \KwTo $K$}{\uIf{$k=1$}
            {$\textbf{z}_0 = \operatorname{E}(\textbf{X}_R)$;}
            \Else{
            $\tilde{\textbf{X}}_G^{k-1} = \operatorname{CT}(\textbf{X}_G^{k-1},\textbf{X}_R)$;
            \tcp{Colour transfer, Eq.(\ref{eq:CT})}
            $\textbf{z}_0 = \operatorname{E}(\tilde{\textbf{X}}_G^{k-1})$;}
            $\textbf{z}_T = \sqrt{\overline{\alpha}_T}\textbf{z}_0 + \sqrt{1-\overline{\alpha}_T} \boldsymbol{\epsilon}, \boldsymbol{\epsilon} \sim \textbf{N (\textbf{0}, \textbf{I})}$;
            \tcp{Diffusion process starts, Eq.(\ref{eq:z_T}) }
            
            \For(\tcp*[h]{$T$ rounds of denoising}){$t=T,\dots,1$}
            {
                
                $\boldsymbol{\epsilon} \sim \textbf{N (\textbf{0}, \textbf{I})}$;
                
                $\hat{\mathbf{z}}_{t-1} = \tilde{\gamma}_t\hat{\mathbf{z}}_t - \tilde{\beta}_t\boldsymbol{\epsilon}_{\boldsymbol{\theta}^*}(\hat{\mathbf{z}}_t, \mathbf{m}, t)+\Tilde{\sigma}_t\boldsymbol{\epsilon}$;
                
                \tcp{Eq.(\ref{eq:infer})}
            }
            $\textbf{X}_G^k = \operatorname{D}(\hat{\textbf{z}}_0)$;

            }
        
\end{algorithm}

\subsubsection{Refinement}
To obtain better image quality, the inference procedure can be repeated based on the synthesized image $\textbf{X}_G$. Specifically, replacing $\textbf{X}_R$ by $\textbf{X}_G$ in Equation~(\ref{eq:z_T}) results in a new starting point $\textbf{z}_{T}^1$ for a new round of progressive denoising (see Equation~(\ref{eq:infer})). Then another  image $\textbf{X}_G^1$ can be synthesized.
This refinement process  continues and stops at $\textbf{X}_G^2$, as illustrated in Figure~\ref{fig:refinement}.
% obtaining

\subsubsection{Color Transfer}
Refinement can improve the generated image quality at the price of reduced similarity to the reference images. Therefore, color-transfer~\cite{reinhard2001color} is used to complement the refinement process. Given the $k$th round refined image $\textbf{X}_G^k$, color-transfer is applied using:
\begin{equation}
    \tilde{\textbf{X}}_G^k = \operatorname{CT}(\textbf{X}_G^k, \textbf{X}_R),
     \label{eq:CT}
\end{equation}
where the style information losses in $\textbf{X}_G^k$ are compensated by color-transfer from $\textbf{X}_R$ directly. The resulting $\tilde{\textbf{X}}_G^k$ can be used for next round refinement, as illustrated in Figure~\ref{fig:refinement-and-color-transfer}.

\subsubsection{Model Ensemble}
% \noindent\textbf{Model Ensemble}\quad 
During model training, multiple parameters can be obtained. We choose two parameters, $\boldsymbol{\theta}^a$ and $\boldsymbol{\theta}^b$, with the best and second best FID performance, respectively, in the validation set. Simple averaging,
\begin{equation}
    \label{eq:model_ensemble}
   \boldsymbol{\theta}^{*} = (\boldsymbol{\theta}^a + \boldsymbol{\theta}^b)/2,
\end{equation}
is then applied. This model ensemble technique can further boost  performance over using $\boldsymbol{\theta}^{*} = \boldsymbol{\theta}^a$.
Algorithm \ref{algthm:infer} shows the complete inferencing process.

\begin{figure*}[t]
    \centering
    \includegraphics[width=1\linewidth]{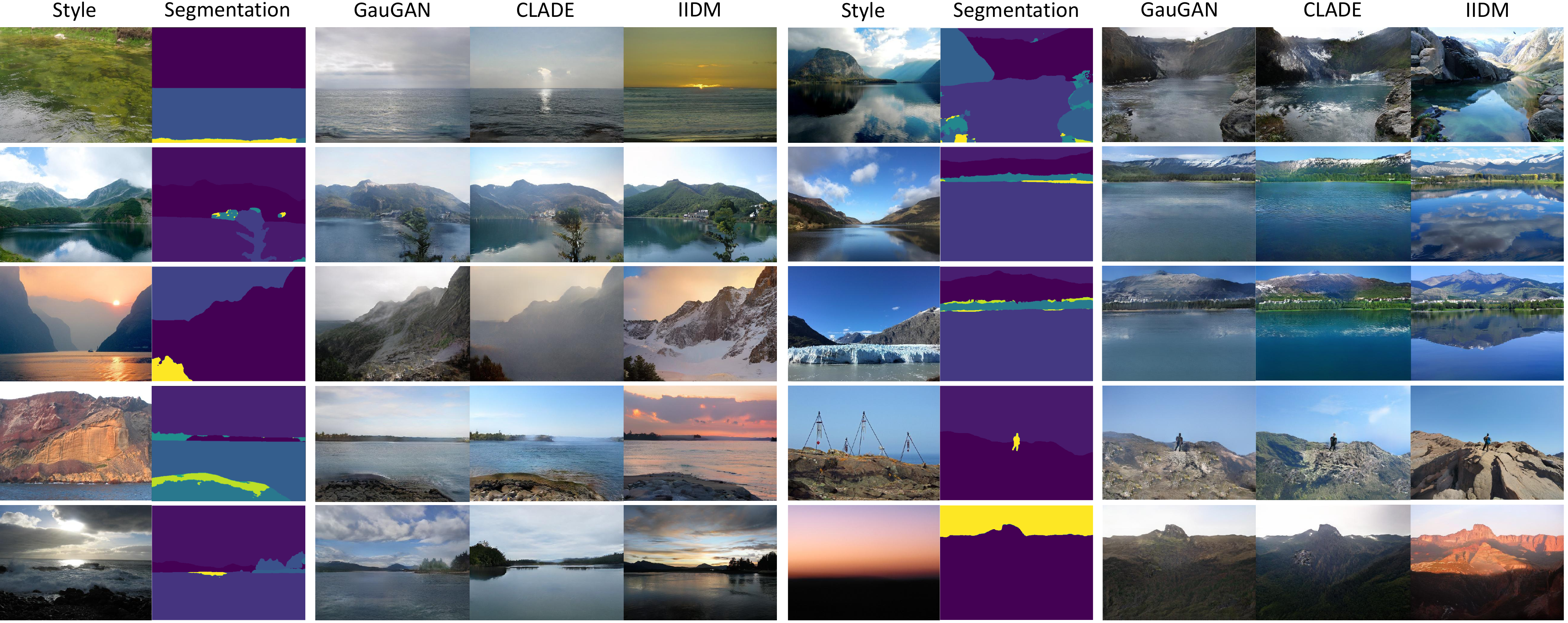}
    \caption{
    Images generated by GauGAN, CLADE, and IIDM, given the same conditional inputs. The outputs of IIDM have higher quality and are more similar in style to the reference image than its counterparts.
    }
    \label{fig:out-compare}
\end{figure*}
\section{Experiments}

\subsection{Dataset}
For the semantic image synthesis track of the Third Jittor Artificial Intelligence Challenge,
$11,000$ high-resolution landscape images (512 pixels width and 384 pixels height) were collected from the Flickr website. Corresponding semantic segmentation maps were also created. Ten thousand image-mask pairs were used for training; the remaining data were used for testing. $1,000$ segmentation masks were provided for testing. Each testing mask corresponded to a reference image from the training set.
Based on the segmentation mask $\textbf{m}$ and the style reference image $\textbf{X}_R$, the synthesized $\textbf{X}_G$ should obey the semantic and style priors.

\begin{table}[t]
\centering
\caption{
Semantic image synthesis metrics for DM, IIDM, and various GAN-based methods. The best performance is in green, and the second best is in red.}
\label{table:main-results}
\begin{tabular}{l|rrrrr}
\hline
Model  
  & $M$↑       & $A$↑     & $FID$↓             & $S$↑          & $T$↑          \\ \hline
SEAN \cite{Zhu_2020_CVPR} & 76.03          & 47.72          & 75.36           & \textcolor{green}{68.38}          & 35.82          \\
SCGAN \cite{wang2021image} & 87.74          & 49.14          & 56.13           & 30.66          & 45.06          \\
GauGAN \cite{park2019semantic} & 89.54          & 49.59          & 43.39           & 54.42          & 54.96          \\
CLADE \cite{tan2021efficient} & \textcolor{red}{92.09}          & \textcolor{green}{50.98}          & 40.63           & 48.14          & \textcolor{red}{57.18}          \\\hline
DM & 91.01         & 49.12          & \textcolor{red}{35.41}           & 27.90          & 55.79          \\
IIDM    & \textcolor{green}{94.15} & \textcolor{red}{50.43} & \textcolor{green}{30.75} & \textcolor{red}{63.76} & \textcolor{green}{65.27} \\ \hline
\end{tabular}
% }
\end{table}

% \noindent\textbf{Metrics}\quad 
Four different metrics were exploited to measure different aspects of the quality of  synthesized images.
Mask accuracy ($M, \%$) quantifies the semantic consistency between the segmentation masks and the synthesized images. The generated images were first segmented by a pretrained SegFormer \cite{xie2021segformer} and then compared to their given masks.
The aesthetic score ($A, \%$) was calculated based on  deep models of aesthetic evaluation~\cite{talebi2018nima}. 
The Fréchet inception distance \cite{heusel2017gans}  (FID $\%$), quantifies the similarity between the distributions of synthesized images and real images.
Style similarity ($S, \%$) is defined as the correlation coefficient between the color histograms of the generated image and of the reference image .
The total score ($T, \%$) is used for  overall assessment and is given by\begin{equation}
    \begin{split}
    T = \frac{M}{100}*[&0.2*A+ 0.3 * (S * 0.5 + 50)+ \\
    & 0.5 * (100-FID)].
    \end{split}
\end{equation}

\begin{figure*}[!t]
    \centering
    \includegraphics[width=0.99\linewidth]{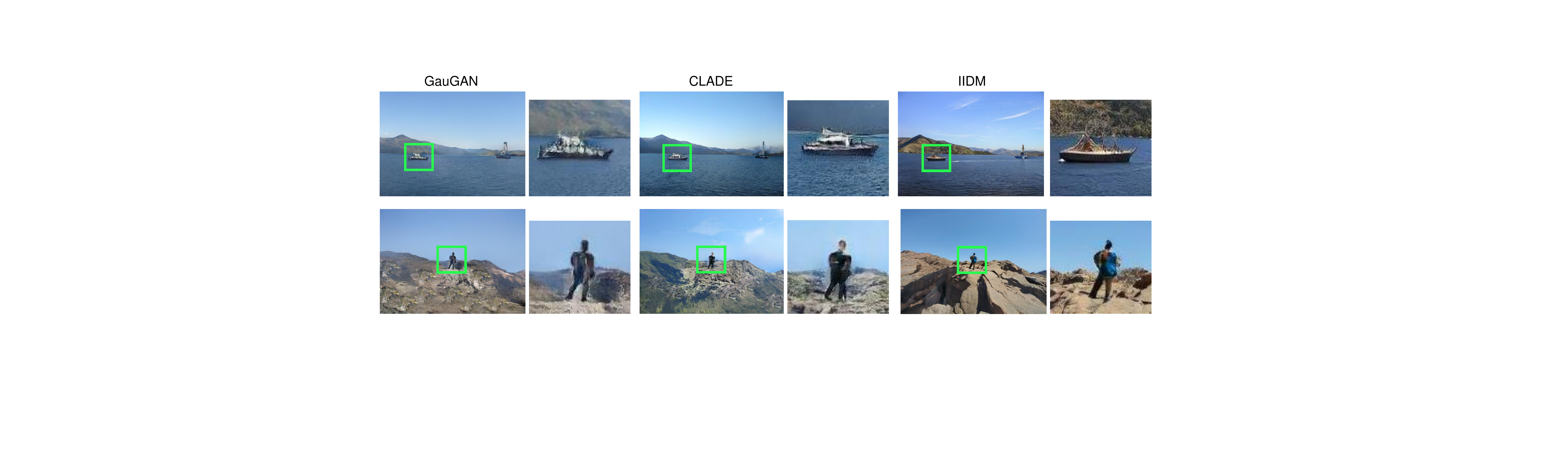}
    \caption{Close-ups of images generated by GauGAN, CLADE, and IIDM. Images generated by IIDM exhibit clearer and more intricate details than the other two.
    % those of its counterparts.}
    }
    \label{fig:details-compare}
\end{figure*}

\begin{figure*}[!t]
    \centering
    \includegraphics[width=0.99\linewidth]{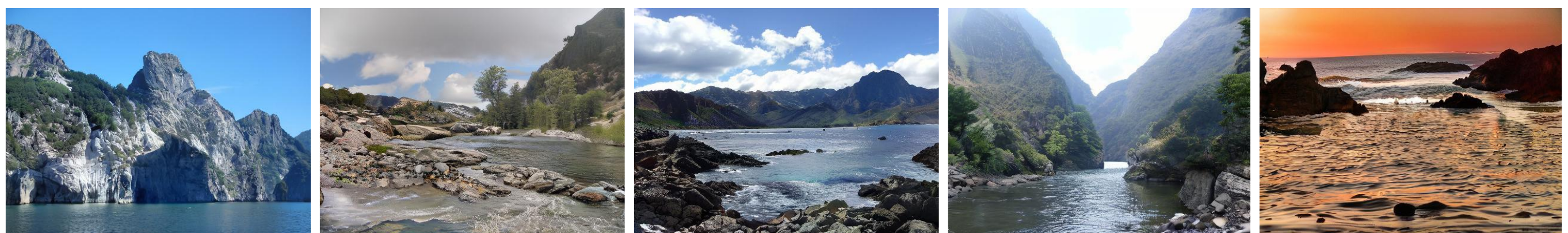}
    \caption{
    Higher resolutions images generated by IIDM.
    % More illustrations of the generations by IIDM with high resolutions.
    }
    \label{fig:more_samples}
\end{figure*}

\begin{table*}[!t]
% \begin{table}[t]
\centering
\caption{
Ablation study for IIDM. 
DM:  basic Diffusion Model in which denoising process starts from  random Gaussian noise.
I2I: Image-to-Image in which denoising  starts from a noise contaminated style reference image.
RF: Refinement used. CT: Color-transfer used. ME: Model ensemble used.
The best performance is in green, and the second best is in red.
}
\label{table:ablation-study}
% \resizebox{0.6\linewidth}{!}{ 
\begin{tblr}
{
  row{1} = {c},row{4} = {c},row{5} = {c},row{6} = {c},
  cell{2}{2} = {c},cell{2}{3} = {c},cell{2}{4} = {c},cell{2}{5} = {c},  cell{2}{6} = {c},  cell{2}{7} = {c},
  cell{2}{8} = {c},  cell{2}{9} = {c},  cell{2}{10} = {c}, cell{3}{1} = {r=4}{},  cell{3}{2} = {c},  cell{3}{3} = {c},  cell{3}{4} = {c},  cell{3}{5} = {c},  cell{3}{6} = {c},  cell{3}{7} = {c},  cell{3}{8} = {c},  cell{3}{9} = {c},  cell{3}{10} = {c},  cell{1}{1} = {c},  cell{1}{2} = {c},  cell{3}{1} = {r=4}{}, 
  vline{2,6} = {-}{},  hline{1-3,7} = {-}{},
  }
                      & I2I  & RF  & CT & ME & $M$↑ & $A$↑ & $FID$↓      & $S$↑ & $T$↑ \\
% CD
DM &     &    &    &    & 91.01    & 49.12      & 35.41  & 27.90  & 55.79 \\
IIDM                & \checkmark   &    &    &    & 93.76    & 49.61      & 32.98 & 62.68 & 63.60  \\
                      & \checkmark   & \checkmark  &    &    & \textcolor{green}{94.57}    & \textcolor{green}{50.43}      & \textcolor{green}{30.63}  & 49.24 & 63.51 \\
                      & \checkmark   & \checkmark  & \checkmark  &    & 94.11    & \textcolor{red}{49.78}      & 31.38 & \textcolor{green}{66.03} & \textcolor{red}{65.09} \\
                      & \checkmark   & \checkmark  & \checkmark  & \checkmark  & \textcolor{red}{94.15}    & \textcolor{green}{50.43}      & \textcolor{red}{30.75} & \textcolor{red}{63.76} & \textcolor{green}{65.27} 
\end{tblr}
% }
\end{table*}

\subsection{Main Results}

% \noindent\textbf{Quantitative Results}\quad
Quantitative results are given in Table~\ref{table:main-results},
our IIDM achieved the best mask accuracy ($94.15\%$) and FID score ($30.75\%$), significantly better than the GAN-based methods.
They demonstrate that IIDM provides superior image quality and  better semantic control than GANs.
Moreover, IIDM achieved the second-best style similarity score $S$ at $63.76\%$,  clearly better than most GAN-based methods.
The progressive generation process of IIDM  flexibly strives for a balance between different conditions. We conclude that IIDM achieves state-of-the-art performance on all criteria.
In comparison, SEAN achieved the highest style similarity score, $68.38\%$, at the price of a reduction in other metrics and thus obtained the worst overall performance. Similar phenomena could also be observed in other GAN-based methods.
The basic diffusion model (DM( generated images with better quality than   GANs, as demonstrated by the superior $FID$ of DM.
However, DM failed to incorporate the style priors, resulting in the worst similarity score.
The proposed IIDM outperformed DM in all metrics.
The overall performance, as measured by total score $T$, of IIDM was better than that of CLADE and DM by clear margins (more than $8\%$).

% Qualitative results
% \noindent\textbf{Qualitative Results}\quad
% Qualitative results are also provided.
Under same conditions of style reference and segmentation map,   images generated by three different models are compared side by side in Figure \ref{fig:out-compare}.
All  methods generated  image contents based on the semantic prior (the segmentation mask).
However, the images generated by IIDM better preserved the style of the reference image than those of competitors.
Close-up views of generated images are also provided in Figure \ref{fig:details-compare}.
The conceptual qualities, e.g., detail richness and boundary sharpness, of the images generated by IIDM are better than the ones from GANs.
In general, such qualitative results are consistent with the quantitative assessments. More  images synthesized by IIDM can be seen in Figure~\ref{fig:more_samples}.

\subsection{Ablation Study} 
% \subsection{Detailed Analysis} 

The effectiveness of different IIDM components was revealed by ab ablation study, with  results in Table~\ref{table:ablation-study}.
The basic Diffusion Model did not incorporate the style reference and thus achieved low style similarity.
IIDM incorporated the style reference as the denoising starting point (see Equation~(\ref{eq:z_T})), interpreted as image-to-image (I2I) translation. I2I significantly improved the style similarity from $27.90\%$ to $62.68\%$.
Using refinement, an inferencing technique, improved the qualities, i.e., mask accuracy, aesthetics, and FID, of generated images at the price of severe style information loss, as low style similarity resulted.
% with low style similarity obtained.
Color-transfer compensated for this loss and substantially increased the style similarity from $49.24\%$ to $66.03\%$ while the positive effects of refinement could still be observed.
Finally, also employing a  model ensemble boosted overall performance $T$ to its best score at $65.27\%$.

\section{Conclusions}
In this paper, we have presented the Image-to-Image Diffusion Model, an efficient and effective semantic image synthesis method that handles both a style reference and a segmentation mask as conditions.
Furthermore, refinement, color-transfer and model ensemble methods can bring substantial improvements at a relatively low price during  inferencing.
% Extensive experiments and analysis are conducted to demonstrate the effectiveness of the proposed method.
The superiority of the proposed method was demonstrated not only by the extensive experiments and analysis conducted but also by winning a challenging competition.

\subsection*{Acknowledgements}
This research was supported by a National Natural Science Foundation for Young Scientists of China award (No. 62106289).

\subsection*{Declaration of competing interest}
The authors have no competing interests to declare that are relevant to the
content of this article.

%%%%%%%%% REFERENCES
% {\small
% \bibliographystyle{ieee_fullname}
% \bibliography{egbib}
% }

\end{document}